\documentclass[10pt,twocolumn,letterpaper]{article}

\usepackage{cvpr}
\usepackage{epsfig}
\usepackage{graphicx}
\usepackage{subfigure}
\usepackage{amsmath}
\usepackage{amssymb}
\usepackage{amsthm}
\usepackage{amsfonts}
\usepackage{mathrsfs}
\usepackage{booktabs}
\usepackage{multirow, eucal}
\usepackage{float}
\usepackage{calligra}

\usepackage[vlined,ruled,linesnumbered]{algorithm2e}

\usepackage{cite}

\usepackage[pagebackref=true,breaklinks=true,letterpaper=true,colorlinks,bookmarks=false]{hyperref}

\usepackage{caption}

\cvprfinalcopy

\begin{document}
	\title{Parallel Attention: A Unified Framework for Visual Object Discovery \\through Dialogs and Queries}

  \author{
  Bohan Zhuang, Qi Wu, Chunhua Shen, Ian Reid, Anton van den Hengel
  \\The University of Adelaide, Australia
  }

	\maketitle

\begin{abstract}
Recognising objects according to a pre-defined fixed set of class labels has been well studied in the Computer Vision. 
There are a great many practical applications where the subjects that may be of interest are not known beforehand, or so easily delineated, however.
In many of these cases natural language dialog is a natural way to specify the subject of interest, and the task achieving this capability (a.k.a, Referring Expression Comprehension) has recently attracted attention.
To this end we propose a unified framework, the ParalleL AttentioN (PLAN) network, to discover the object in an image that is being referred to in variable length natural expression descriptions, from short phrases query to long multi-round dialogs. The PLAN network has two attention mechanisms that 
relate parts of the expressions 
to both the global visual content and also directly to object candidates. Furthermore, the attention mechanisms are recurrent, making the referring process visualizable and explainable. The attended information from these dual sources are combined to reason about the referred object. These two attention mechanisms can be trained in parallel and we find the combined system outperforms the state-of-art on several benchmarked datasets with different length language input, such as RefCOCO, RefCOCO+ and GuessWhat?!. 

\end{abstract}

	\section{Introduction}

Despite the fact that Object Detection has become a figurehead challenge in Computer Vision, the number of applications of the technology is limited by the fact that it demands a pre-defined set of class labels and large number of pre-prepared training images for each.  This means not only that the list of objects of interest is fixed, and determined long in advance, but also that their appearance must remain fixed.
The limitations of this approach are visible in the rise of mediation strategies such as Domain Adaptation~\cite{Lu_2017_ICCV, gong2016domain}, Transfer Learning~\cite{raina2007self, patricia2014learning}, Zero-Shot learning~\cite{qiao2016less, Zhang_2017_CVPR}, and a subset of the Meta-Learning approaches\cite{santoro2016meta}.
More fundamentally, the traditional Machine Learning approach whereby the problem and training data are specified before the solution is devised, and then trained, inherently implies that every instance of the target class is of interest, in every image it processes.

\begin{figure}
\begin{center}
	\resizebox{0.97\linewidth}{!}
	{
		\begin{tabular}{c}
			\includegraphics{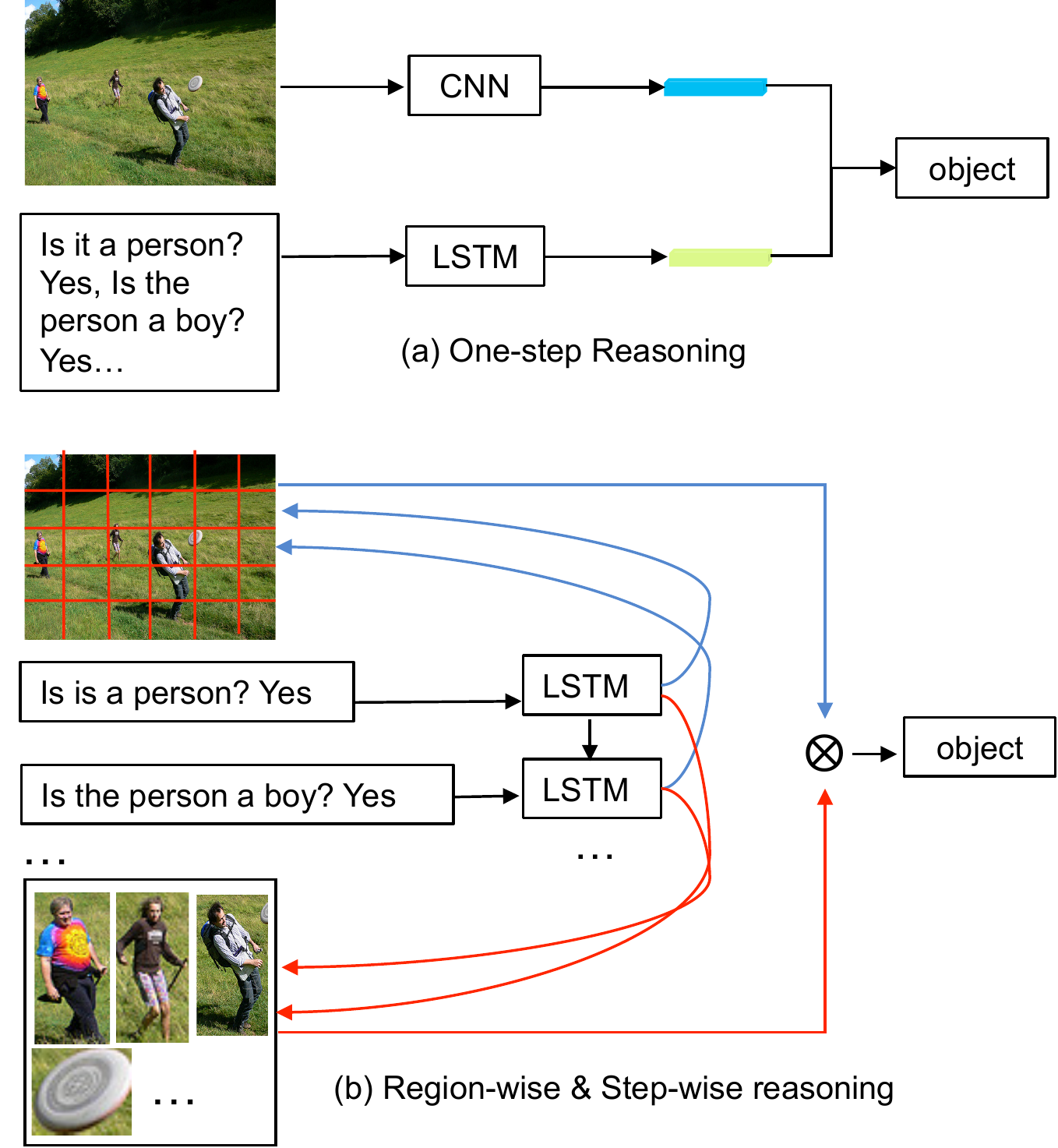}
		\end{tabular}
	}
\end{center}
\vspace{-12pt}
   \caption{One-step reasoning vs. our proposed region-wise\&step-wise reasoning framework for the task of referring expression. In conventional frameworks (a), visual and language features are embedded into a joint space for one-step reasoning. However, in our approach (b), we propose to recurrently discover the target with a parallel attention mechanism, with region and step-wised query.}
   \vspace{-12pt}
\label{fig:short}
\end{figure}

A more flexible, and practically applicable approach is to define the subject of interest at test time, not least because it allows instance-level detection.  It also supports the specification of subjects of interest that were unimaginable at training time.  Better even than a system that allows test-time specification of the subject of interest, however, is one that will interactively cooperate towards an accurate detection, particularly in the presence of ambiguity.
This is the challenge we consider here.

Referring expressions are used frequently by people to identify or indicate particular objects within their physical environment \cite{Yu_2017_CVPR}. The length of the expressions can range from a very short phase (2-3 words) to a multi-round dialog (such as the GuessWhat?! game). The longer the expression is, the more information is provided, however, the harder the problem is because more details need to be analysed and more steps of reasoning are required. Previous works \cite{yu2016modeling,Yu_2017_CVPR,Vries_2017_CVPR} have primarily appled a CNN-RNN pipeline that uses a CNN to encode the image content and an LSTM to encode the expression. 
The encoded features are then jointly embedded and used to
to locate the object that is related to the expression, as shown in the Figure \ref{fig:short} (a). This mechanism works well when the expression is short and the number of the proposed objects is limited. However, when the expression is too long, in dialog form, or there are too many potential objects in the image, using the global representation to encode the expression and image into a single vector 
fails because such a one-step process does not have the flexibility required to relate multiple parts of the expression to multiple parts of the image.

Instead, we argue that the representation of expression should be stepwise and the representation of image should be region-wise. More significantly, we argue that the referring process should be also carried out stepwise, \ie after a stepwise representation (such as a word in a sentence or a question-answer pair in a dialog) is given, it should first focus on some potential regions, with the expression or dialog being `listened to' continually, the number of potential regions decreases and the area of potential interest becomes smaller, until it converges to a single region, as shown in the figure \ref{fig:short} (b). For example, given an expression `the woman in the middle wearing brown jacket', regions corresponding to all of the women in the image are considered first. After reading `in the middle', regions of `woman' at the border of the image are eliminated. Finally, once the input `wearing brown jacket' is given, only one region is left. 

On the basis of the above we propose a unified framework, the ParalleL AttentioN (PLAN) network, which is the main contribution of this paper, to recurrently discover the object in an image that is being referred to  natural language descriptions, from short phrases query to long multi-round dialogs. Specifically, a two-way attention mechanism is applied to localize the referred-to region in the global contextual features from the whole image (\ie, image-level attention) and to select the referred-to proposal region (\ie, proposal-level attention) from a set of such region proposals.  
The `image-level' attention can help identify referring expressions that are related to the contextual information, such as `the girl outside the cinema'. The `cinema' here is then the global scene context we need to consider in the reasoning process. The `proposal-level' attention learns to weight different candidate object proposals, when the referring expression is associated with multiple objects, such as `the girl at the left side of the fire hydrant', then the model will learn to focus on the object proposal of `girl' and the `fire hydrant', but not the global scene. The attention mechanism proposed here allows the model to sequentially `listen to' the referring expression so that a step-wise reasoning process is achieved. We evaluate our PLAN model on what is currently the largest referring expression dataset, the ReferCOCO, ReferCOCO+ and ReferCOCOg \cite{yu2016modeling}, which has a different length of the input expression. Our model outperforms the previous state-of-art by a large margin, for example, our single model even outperforms an ensemble model (and with Reinforcement Learning) on a test split on the ReferCOCO. We further evaluate our model on the recently released GuessWhat?!\cite{Vries_2017_CVPR} dataset, which requires an agent to point out the object in an image that is being discussed by a `Questioner' and an `Oracle' via multiple rounds of dialog. Here our model also outperforms the previous state-of-art significantly. 

As a side contribution, because our model uses a recurrent process to discover the object, we can display the referring updating process along with the input expression query and the dialog, which makes the multiple steps of reasoning in the Refer Expression Comprehension visualizable and explainable. Qualitative results (see Figure \ref{fig:exp_guesswhat}) show that we can produce excellent visualization performance.

	\section{Related work}

\paragraph{Referring Expressions}
The are two distinct challenges relating to Referring Expressions: generation and comprehension. The generation task requires a model to generate a language expression for the given region, which is very similar to the dense image captioning task\cite{johnson2016densecap}. Referring expression comprehension aims to localize the regions being described by a given referring expression~\cite{yu2016modeling, nagaraja2016modeling, mao2016generation, hu2016natural, Hu_2017_CVPR}. Given a set of extracted candidate regions, each region is scored by the model with respect to the referring expression and the region with the highest score is selected as the final grounding result. In this paper, we mainly focus on improving the referring expression comprehension task rather than the generation task.

\begin{figure*}[t!]
	\centering
	\resizebox{1\linewidth}{!}
	{
		\begin{tabular}{c}
			\includegraphics{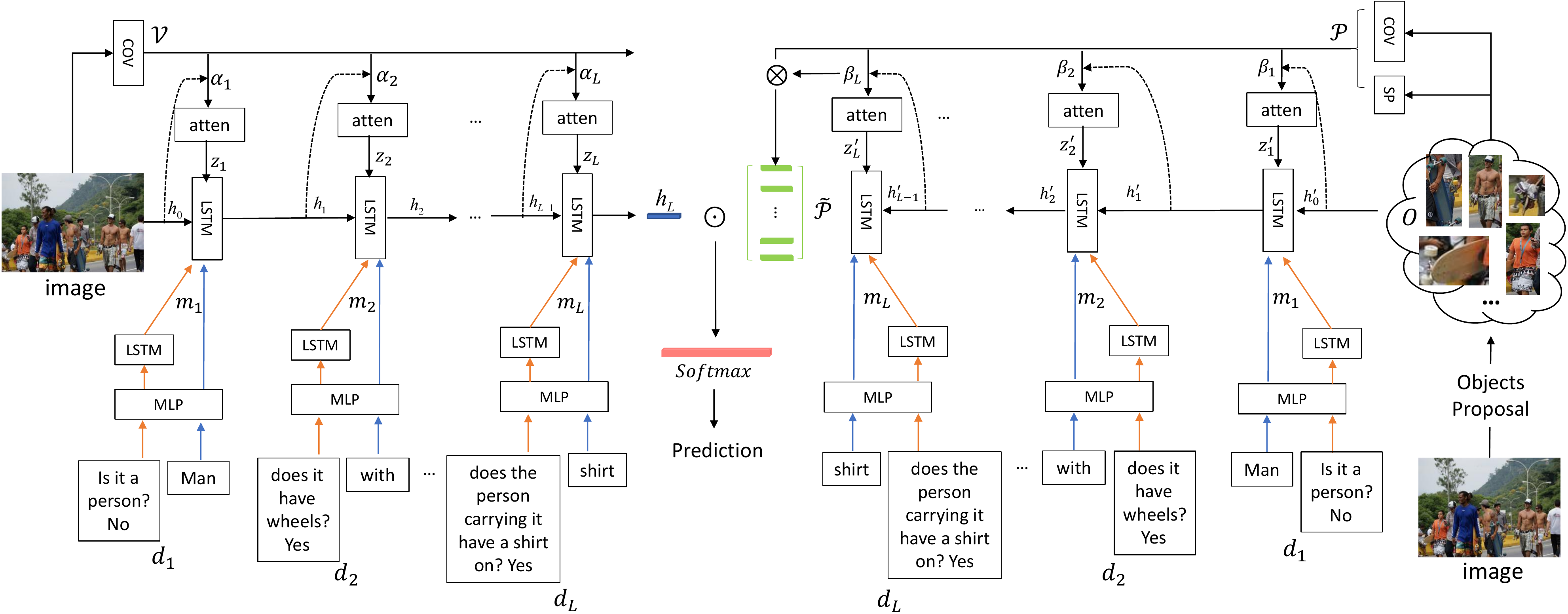}
		\end{tabular}
	}
	\vspace{-5pt}
	\caption{The proposed ParalleL AttentioN (PLAN) framework for recurrent object discovery. The left-side is the image-level framework that is used to encode the entire image information and the referring language descriptions by recurrently attending on different image regions. The right side is the proposal-level framework with the proposal regions and referring expressions as input. It is designed to recurrently attend on proposals based on the language information and mutually encode both visual and textual representations. The both side representations are further combined to effectively localize the referring target. An additional LSTM encoder (orange arrow) is added if the expression unit $d_i$ is a Question-Answer pair, otherwise, we directly input the word into a MLP, followed by an LSTM.}
	\label{fig:framework}
	\vspace{-8pt}
\end{figure*}

Several works~\cite{mao2016generation, hu2016natural} propose to use local visual features or global image features as the feature representation. To better employ the structural information between different candidates, some works~\cite{nagaraja2016modeling, yu2016modeling, Hu_2017_CVPR} further explicitly incorporating modeling context between objects into referring expression. In~\cite{Hu_2017_CVPR, nagaraja2016modeling}, the models are proposed to handle inter-object relationships for grounding a referential expression into a pair of regions. Yu \etal~\cite{yu2016modeling} argues that visual appearance comparison can help localize the referred object by looking into other surrounding candidates. Luo \etal \cite{luo2017comprehension} propose to utilize models trained for comprehension task to generate better expressions. Different from these methods, we propose to incorporate both contextual image features and local visual features in a unified framework, the both features are weighted by a parallel attention mechanism, which make us recurrently discover the object that referred by the expression. 

\vspace{-13pt}
\paragraph{Phrase grounding}
Phrase grounding aims to localize the objects described in the phrase. The main problem is to learn the correlation between visual and language descriptions. To solve this problem, Karpathy \etal ~\cite{karpathy2014deep} propose to align the objects and the fragments of language into the embedding space with a structured max-margin objective and~\cite{karpathy2015deep} further replace the dependency tree of the language parser with a bidirectional RNN. Some approaches~\cite{plummer2015flickr30k, wang2016structured} use a Canonical Correlation Analysis~\cite{hardoon2004canonical} based method to learn the correlation between visual and language modalities. 
Recently, Hu \etal~\cite{hu2016natural} proposed an SCRC model to integrate spatial configuration and global scene-level contextual information into the network.

\vspace{-13pt}
\paragraph{The Attention Mechanism}
An attention mechanism was first successfully introduced within the image captioning task by~\cite{xu2015show}. Based on this, Lu \etal~\cite{lu2016hierarchical} further proposed a co-attention model for VQA that jointly reasons about language and the image. Then in~\cite{Lu_2017_CVPR}, an adaptive attention model with a visual sentinel was applied to decide when to attend. You \etal \cite{you2016image} run a set of attribute detectors to get a list of visual attributes and fuse them into the RNN hidden state. Rather than statically attending to the image using the given expressions, we instead propose to recurrently update the attention weight with the stepwise expression query or the dialog. Also we propose a parallel attention mechanism for embedding global and local features in a unified framework. 

\vspace{-13pt}
\paragraph{Vision and Language}
Our work is part of a group of recent methods combining vision and language. Recent work in this area includes a variety of approaches to image captioning~\cite{xu2015show, yang2016review, johnson2016densecap, vinyals2015show,wu2016value, Krishna_2017_ICCV, Liu_2017_ICCV, Li_2017_ICCV, Pedersoli_2017_ICCV, Gu_2017_ICCV} and visual question answering (VQA)~\cite{lu2016hierarchical, wu2016ask, Hu_2017_ICCV, Zhu_2017_ICCV, Yu_2017_ICCV, Gan_2017_ICCV, Kafle_2017_ICCV, Ben-younes_2017_ICCV}. Image captioning seeks to generate a natural language description for the whole given image, while the VQA requires the agent to answer previously unseen visual questions about an image. Most recently, a new vision-and-language task, Visual Dialog \cite{Das_2017_CVPR, Vries_2017_CVPR, Das_2017_ICCV}, demands that an agent participate intelligently in a dialog about an image. In our work, we extend the phase/sentence based referring expression task to dialogs, \ie, extending the dialog task proposed in~\cite{Vries_2017_CVPR} to the problem of identifying a specific object in an image.

	 \section{The PLAN Model} \label{sec:method}

In this section, we describe our unified model that takes as input an image and a set of natural language expressions and outputs a bounding box that contains the object that is referred by the expressions. Formally, we denote an input image as ${I}$ containing a set of $N$ object proposals $\mathcal{O}=\{ {{{O}}_1},{{{O}}_2},...,{{{O}}_N}\}$. For the GuessWhat benchark only, each object candidate is assigned an object category ${c_i} \in \{ 1,...,C\}$ 
where $C$ is the number of object categories. 
The set of referring expressions is $\mathcal{D} = \{ {{{d}}_1},...{{{d}}_L}\}$, where 
${d_i}$ is either a question-answer pair or a single word, depending on whether the input is dialog or a sentence description, and L is total number of expressions.
The output of the proposed model is a probability distribution over the object proposals and the region with the highest probability is selected as the grounding result. 

The core of our proposed model relies on a ParalleL AttentioN (PLAN) model, which takes the encoded image features, language expressions features and the candidate proposal object features as  input. On one side, we use word level (for phase and sentence input) or question-answer pair level (for dialog input) features to guide the image attention, \ie, to attend to the regions that are correlated with the expression. At the other side, we use the language features to recurrently weight the object proposasl until all the expressions have been `listened to'. The parallel attended features are combined to reason about the target object. 

In the following sections, we first describe the feature encoding for the different inputs in Sec.~\ref{sec:recurrent} and explain how to recurrently discover the target object with parallel attention mechanism in Sec.~\ref{sec:attention}. The implementation details are described in Sec.~\ref{sec:implementation} and the whole framework is illustrated in Figure~\ref{fig:framework}. 

\subsection{Feature encoding} \label{sec:recurrent}

\paragraph{The global visual feature}\label{sec:visual}
To encode the full image while maintaining the spatial information therein, we first rescale the image to 224x224 and then pass it through a VGG-16 network \cite{simonyan2014very} pre-trained on ImageNet to obtain its $Conv5\_3$ feature, denoted as ${\mathcal{V}} = \{{{\bf{v}}_1},...,{{\bf{v}}_K}\}$, where $K = 49$.%

\vspace{-12pt}
\paragraph{Object proposal features}
For each candidate object proposal $O_i$, the corresponding feature is composed of two parts, the first is the CNN feature $u_i$ extracted as described for the global visual features, the $Conv5\_3$ feature from the VGG-16 for each object proposal. To further increase the expressive power, similar to~\cite{hu2016natural, Vries_2017_CVPR, Yu_2017_CVPR, yu2016modeling,Hu_2017_CVPR, mao2016generation}, we also embed the spatial representations of the proposal regions. Following~\cite{Yu_2017_CVPR, Vries_2017_CVPR}, the spatial information of the bounding box of each object is encoded as the an 8-d vector:
${s_i} = [ {x_{\min }},{y_{\min }},{x_{\max }},{y_{\max }},{x_{center}},{y_{center}},{w_{box}},{h_{box}}]$, 
where $w_{box}$ and $h_{box}$ are width and height of the bounding box. The image height and width are normalized to the range $[-1, 1]$ and the center of the image is set as the origin. For the experiments on the GuessWhat?! \cite{Vries_2017_CVPR} dataset, the additional object category information for each candidate is also used, for a fair comparison with the previous state-off-the-art. We denote the feature representation as ${\mathcal{P}} = \{{{\bf{p}}_1},...,{{\bf{p}}_N}\}$, where $N$ is the number of object proposals and each ${\bf{p}}_i=[u_i;s_i;c_i]$, where $c_i$ is the object category label for the $i$-th object\footnote{$c_i$ is only used in GuessWhat?! experiment for a fair comparison. We additionally evaluate the model without this object category information.}.

\vspace{-12pt}
\paragraph{Referring expression features} \label{sec:expressions}

Since we argue that the representation of the expression should be processed stepwise, 
in contrast to previous work that encodes the whole expression into a single vector, we represent each expression unit ${{d}}_i$ separately. Each ${{d}}_i$, is first tokenized by the word tokenizer from the nltk toolkit~\cite{bird2009natural}. Each word is represented as an one-hot vector in a pre-constructed dictionary $\mathbb{D}$. Then we convert each one-hot vector into a dense embedding with a learnable MLP layer. If the expression is a dialog (multiple rounds of question-answer pairs), we pass the expression unit ${{d}}_i$ (a QA pair as a whole) to an LSTM encoder and take the last hidden state as the representation (as shown in Figure \ref{fig:framework} by the orange arrow). If the expression is a single sentence, the output of the MLP for each word will be directly used as the representation for the expression unit ${{d}}_i$ (as shown in the Figure \ref{fig:framework} by the blue arrow). For simplicity, we define $ {\mathcal{M}} = \{ {{\bf{m}}_1},...,{{\bf{m}}_L}\}$ as the encoded representation for the expression unit ${{d}}_i$, for either case.

\subsection{Recurrent parallel attention mechanism}\label{sec:attention}
Given the encoded visual feature representations and a sequence of encoded language descriptions, the next step is to utilize these effectively to identify the correct target object. One naive approach is to concatenate all the features as a vector representation and then dot-product with the features of candidate objects one-by-one, followed by a softmax to obtain a distribution over the objects. The object with the highest probability will be treated as the referring target. However, in this approach, the language and the visual features are encoded separately without learning from each other. Ideally, the regions that are concentrated on should be updated along with the referring expression being `listened to'. To solve this problem, we propose to recurrently discover the object by sequentially `listening to' the referring expressions. What's more, a parallel attention mechanism is applied to localize the referring region within the global contextual features from the whole image (\ie image-level) and to select the referring proposal (\ie proposal-level).

\subsubsection{Image-level attention}\label{sec:image-level}
Image-level attention corresponds to the left portion of Figure~\ref{fig:framework}, within which we use expression features to attend over the global image. We use an LSTM to encode the visual feature and the language feature simultaneously. We define ${{\bf{h}}_t}$ as the hidden state at time $t$ as:
\begin{equation}
{{\bf{h}}_t} = LSTM({{\bf{m}}_t},{{\bf{z}}_t},{{\bf{h}}_{t - 1}})
\end{equation}
where ${{\bf{m}}_t}$ is the feature representation of the $t$-th expression unit (that is ${\bf{m}}_i$ above, where $i=t$), ${{\bf{z}}_t}$ is the attended image features at time $t$. Note that the index of the language description corresponds to the time step in LSTM encoding. 
We compute ${{\bf{z}}_t}$  with an attention mechanism,  
with the convolutional image feature ${\mathcal{V}} = \{{{\bf{v}}_1},...,{{\bf{v}}_K}\}$ and ${{\bf{h}}_{t - 1}}$, which is the hidden state of the LSTM at time $t-1$. We feed them through a MLP layer separately followed by a softmax function to get the attention distribution over the spatial location on $\mathcal{V}$:
\begin{equation}
\begin{array}{l}
{{\bf{e}}_{ti}} = \tanh {({{\bf{W}}_v}{\bf{v}}_i}+{{\bf{W}}_h}{{\bf{h}}_{t - 1}}),\\
{{\boldsymbol{\alpha}}_{ti}} = softmax ({{\bf{e}}_{ti}}),
\end{array}
\end{equation}
where ${{\bf{W}}_v}$ and ${{\bf{W}}_h}$ are the embedding matrices and ${{\boldsymbol{\alpha}}_{ti}} $ is the attention weights over the convolutional feature ${\bf{v}}_i$ at time step $t$. Then the attended image feature can be obtained by:
\begin{equation}
\vspace{-2pt}
{{\bf{z}}_t} = \sum\limits_{i = 1}^K {{\boldsymbol\alpha _{ti}}} {{\bf{v}}_i}
\vspace{-2pt}
\end{equation}
We thus use the previously encoded language and image information ${{\bf{h}}_{t - 1}}$ as context to attend on the image. And, in return, we utilize the attended image feature ${{\bf{z}}_t}$ as context together with the language representation ${{\bf{m}}_t}$ to predict the next hidden state of the LSTM. With this strategy,  we can recurrently encode different image information given different referring language descriptions, and the last hidden state (${\bf{h}}_{t=L}$) of the LSTM is taken as the final representation of the image-level attention, which will be used in the final referring step.

\vspace{-5pt}
\subsubsection{Proposal-level attention} \label{sec:proposal-level}
The proposal-level attention corresponds to the right-hand side of Figure~\ref{fig:framework}. In contrast to the image-level attention that applies over the densely divided regions, the proposal-level attention model focuses on only the proposal regions that contain the candidate objects. The input is thus a set of proposal regions and a sequence of language descriptions, and the model should learn to recurrently attend to different candidates with varying language description input. Similar to the image-level attention, the hidden state at time $t$ can be represented as:
\begin{equation}
{\bf{h}}_t^{'} = LSTM({\bf{m}}_t,{\bf{z}}_t^{'},{\bf{h}}_{t - 1}^{'})
\end{equation}
where ${\bf{z}}_t^{'}$ is the summation of the attended proposal features. Given the last hidden state ${\bf{h}}_{t - 1}^{'}$ of the LSTM and the proposal container ${\bf{P}} = \{ {{\bf{p}}_1},...,{{\bf{p}}_N}\}$, we derive the attention score for each proposal as follows:
\begin{equation}
\begin{array}{l}
{\bf{e}}_{ti}^{'} = \tanh {({\bf{W}}_p{{\bf{p}}_i}}+{\bf{W}}_h^{'}{\bf{h}}_{t - 1}^{'}),\\
{\boldsymbol{\beta}_{ti}} = softmax( {{\bf{e}}_{ti}^{'}}) ,
\end{array}
\end{equation}
where ${\boldsymbol\beta _{ti}}$ is the attention score for proposal ${{\bf{p}}_i}$ at time $t$, and ${\bf{z}}_{t}^{'}$ is computed by:
\begin{equation}
\vspace{-2pt}
{{\bf{z}}_t^{'}} = \sum\limits_{i = 1}^N {{\boldsymbol\beta _{ti}}} {{\bf{p}}_i}
\vspace{-2pt}
\end{equation}

In contrast to the image-level attention that uses the final hidden state to represent the information, for the proposal-level attention, we use the learned proposal attention weights at the last time step (when the last expression $d_L$ is `listened to') $\boldsymbol\beta_{Li}$ to weight each of the proposal candidate objects features ${\bf{p}}_i$, to obtain the attended features for each object proposal:
\begin{equation}
\widetilde{\bf{p}}_i=\boldsymbol\beta_{Li}{\bf{p}}_i
\end{equation}
The output of the proposal-level attention is thus the $\widetilde{\mathcal{P}}=\{\widetilde{\bf{p}}_1,\cdots,\widetilde{\bf{p}}_N\}$, where $N$ is the number of proposals.

\subsubsection{Referring process}
After obtaining the final representations of the two-level attentions, we then need to fuse them to localize the final target. Intuitively, given the same referring language descriptions, the model should encode the similar visual information and the language representation. Thus, we just use a simple dot-product followed by a softmax function to obtain the prediction distribution over the proposals:
\begin{equation}
P_i=softmax({\bf{h}}_{t=L}\odot\widetilde{\bf{p}}_i)
\end{equation}
where $P_i$ 
is the probability that expression $D$ is referring to object $O_i$,
 and $\odot$ is the dot product. Finally, we use the cross-entropy loss as the objective.

\subsection{Implementation details} \label{sec:implementation}

We optimize our model using Adam~\cite{kingma2014adam} in Pytorch. The learning rate is initialized to 0.001 and divided by 10 after 15 epochs with batch size of 32. The hidden state size of the LSTM is set to 512. We also set the embedding dimension for each MLP layer to 512. To avoid over-fitting, we add dropout with a ratio of 0.4 after each linear transformation in the MLP layers.

	 \section{Experiments}

\begin{table*}[t!]
	\centering
	\resizebox{0.87\linewidth}{!}{
		\begin{tabular}{ l | c c c | c c c | c }
			\hline
			\multirow{2}{*}{Method} & \multicolumn{3}{c|}{RefCOCO} & \multicolumn{3}{c|}{RefCOCO+} &RefCOCOg \\\cline{2-8}
			&val &TestA &TestB  &val  &TestA &TestB  &val\\\hline
			MMI~\cite{nagaraja2016modeling}  &- &71.72\% &71.09\%  &- &58.42\% &51.23\% &62.14\%\\
			visdif~\cite{yu2016modeling} &- &67.57\% &71.19\% &-  &52.44\% &47.51\%  &59.25\%\\
			visdif+ MMI~\cite{yu2016modeling} &- &73.98\% &76.59\% &- &59.17\% &55.62\% &64.02\% \\
			Neg Bag~\cite{nagaraja2016modeling}  &- &75.60\% &78.00\% &- &- &- &68.40\% \\
			Luo \etal~\cite{luo2017comprehension}  &- &74.14\% &71.46\% &- &59.87\% &54.35\% &63.39\% \\
			Luo \etal (w2v)~\cite{luo2017comprehension} &- &74.04\% &73.43\% &- &60.26\% &55.03\% &65.36\%\\\hline
 			listener~\cite{Yu_2017_CVPR}  &77.48\% &76.58\% &78.94\% &60.50\% &61.39\%  &58.11\%  &71.12\% \\
 			speaker+listener~\cite{Yu_2017_CVPR} &77.84\% &77.50\% &79.31\% &60.97\% &62.85\% &58.58\% &72.25\%\\
 			speaker+listener+reinforcer~\cite{Yu_2017_CVPR} &78.14\% &76.91\%&80.10\% &61.34\% &63.34\% &58.42\% &71.72\% \\ 
 			speaker + listener + reinforcer (ensemble) ~\cite{Yu_2017_CVPR} &78.88\% &78.01\% &80.65\% &61.90\% &64.02\% &59.19\% &\bf{72.43}\% \\\hline
			Baseline &78.16\%   &77.45\%  &79.54\%  & 62.41\%  &63.48\%   &59.30\%  &66.05\%  \\
			Image-level attention&79.20\%  &78.29\%  &80.11 \% &63.27 \%  &64.16\%  &60.13\% & 67.89\% \\
			Proposal-level attention & 81.09\% &80.13\% &80.84\% &63.57\% &65.53\% &60.52\% &68.54\% \\
			ParalleL AttentioN (PLAN) &\bf{81.67}\% &\bf{80.81}\% &\bf{81.32}\% &\bf{64.18}\%  &\bf{66.31} \%  &\bf{61.46} \%  &69.47\%\\\hline
		\end{tabular}}
		\vspace{-2pt}
		\caption{Accuracies on RefCOCO, RefCOCO+ and RefCOCOg datasets. Note that for RefCOCOg, we use the standard google split for testing while ~\cite{Yu_2017_CVPR} divide the dataset by randomly partitioning objects into training and validation splits, which is not comparable.}
		\vspace{-10pt}
		\label{tab:refcoco}
	\end{table*}

To demonstrate the benefits of each module, we analyse the effect of the recurrent encoding baseline, image-level attention, proposal-level attention and the full model. Four methods are implemented and compared:
\begin{enumerate}
	\vspace{-3pt}
	\item ``\textbf{Baseline}'': %
	Different from the full model, the second-level LSTM only takes the embeddings of the language descriptions $\mathcal{M}$ as input without any attention mechanism. We then concatenate the last hidden state ${{\bf{h}}_L}$ and the extracted image feature $\mathcal{V}$ to obtain the image-level representation. For the proposal-level representation, we concatenate the visual representations for proposals and their corresponding spatial feature together. We still use dot-product to calculate the final scores over the objects. 
	\vspace{-5pt}
	\item ``\textbf{Image-level attention}'': Based on the baseline model, we further add the recurrent image-level attention as described in Sec~\ref{sec:image-level}. 
	\vspace{-5pt}
	\item ``\textbf{Proposal-level attention}'': Based on the baseline model, we further add the recurrent proposal-level attention as described in Sec~\ref{sec:proposal-level}, but without using the image-level attention.
	\vspace{-5pt}
   \item ``\textbf{Parallel attention}'': We implement our full PLAN model based on Sec~\ref{sec:method}.
\end{enumerate}
In addition, we also compare the performance of our methods against those reported in the related literature.

\subsection{Datasets} \label{sec:dataset}
We evaluate the performance on four referring expression datasets, including RefCOCO, RefCOCO+, RefCOCOg~\cite{yu2016modeling} and GuessWhat?!~\cite{Vries_2017_CVPR}. All of the datasets are collected on MS-COCO images~\cite{lin2014microsoft}. 

\begin{figure*}[t!]
	\centering
	\resizebox{0.9\linewidth}{!}
	{
		\begin{tabular}{c}
			\includegraphics[width=50cm]{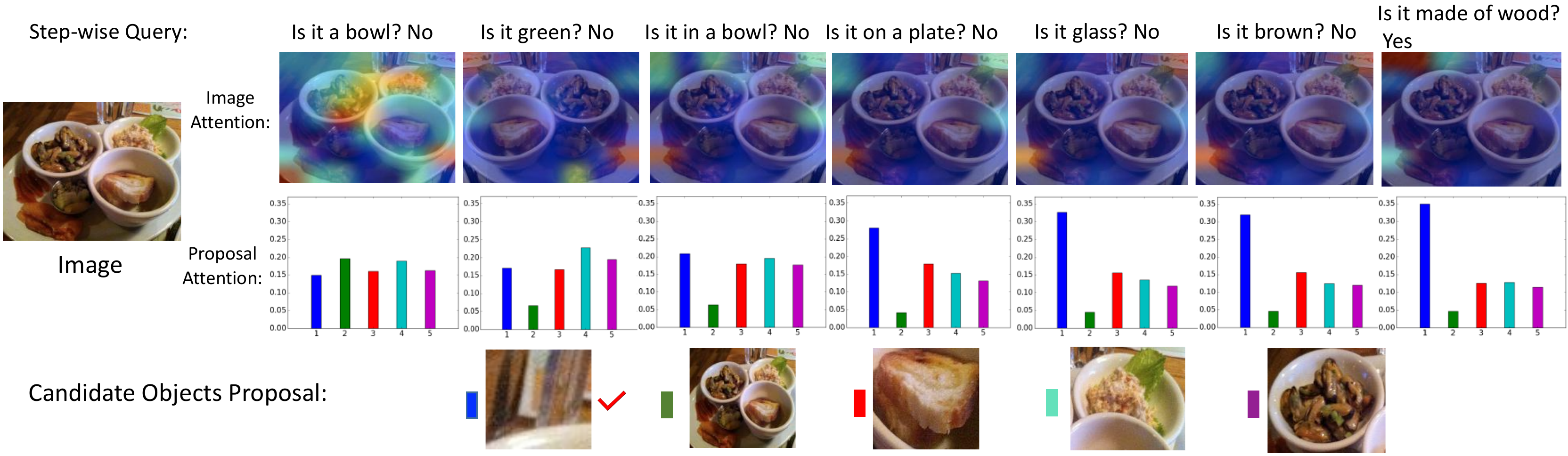} \vspace{1pt}\\\hline \vspace{1pt}
			\includegraphics[width=50cm]{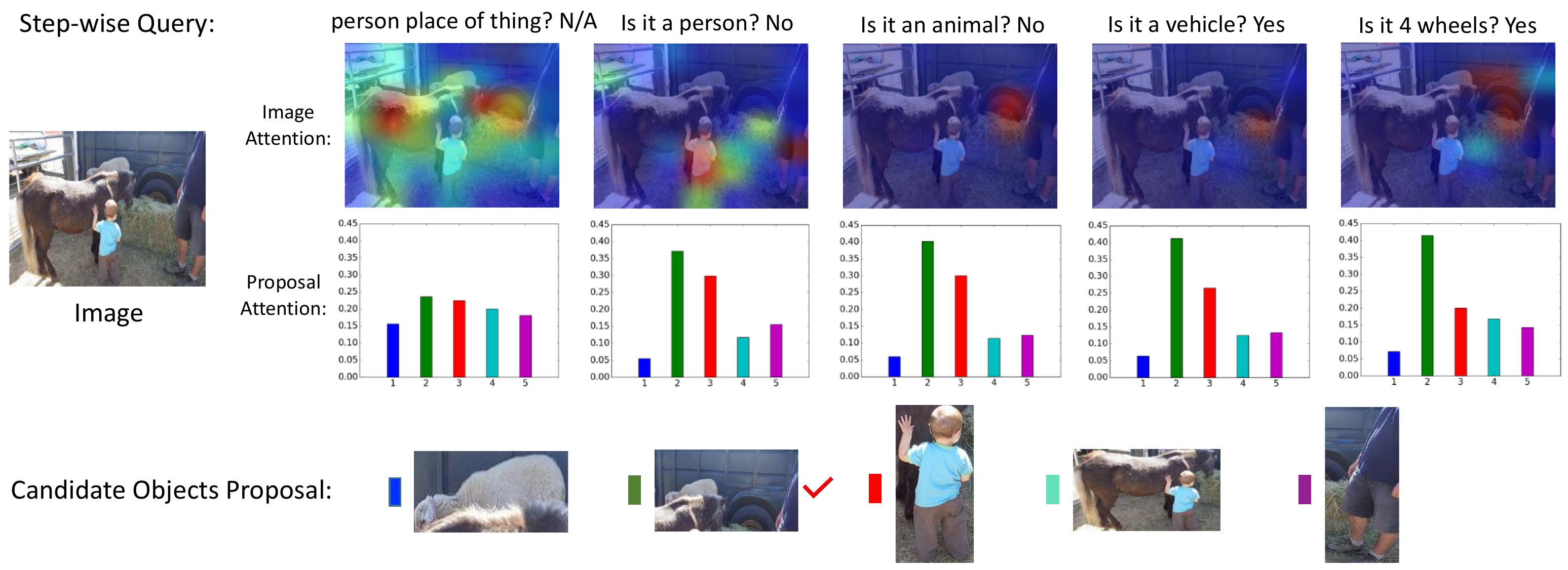} \vspace{1pt}\\\hline \vspace{1pt}
			\includegraphics[width=50cm]{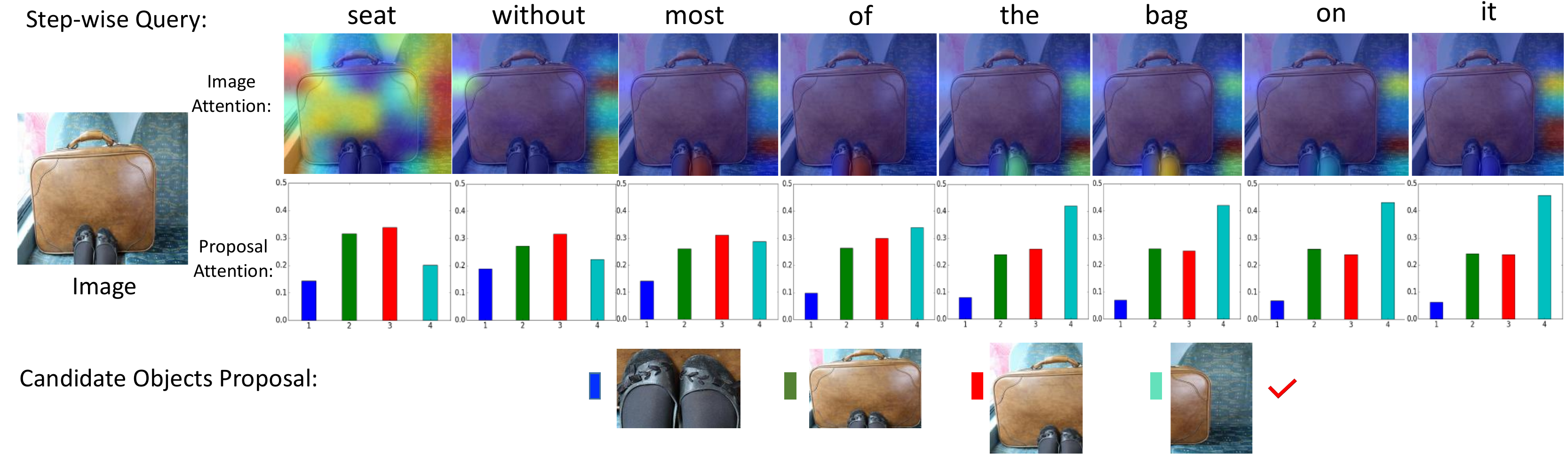} \vspace{1pt}\\\hline\hline \vspace{1pt}
			\includegraphics[height=20cm,width=50cm]{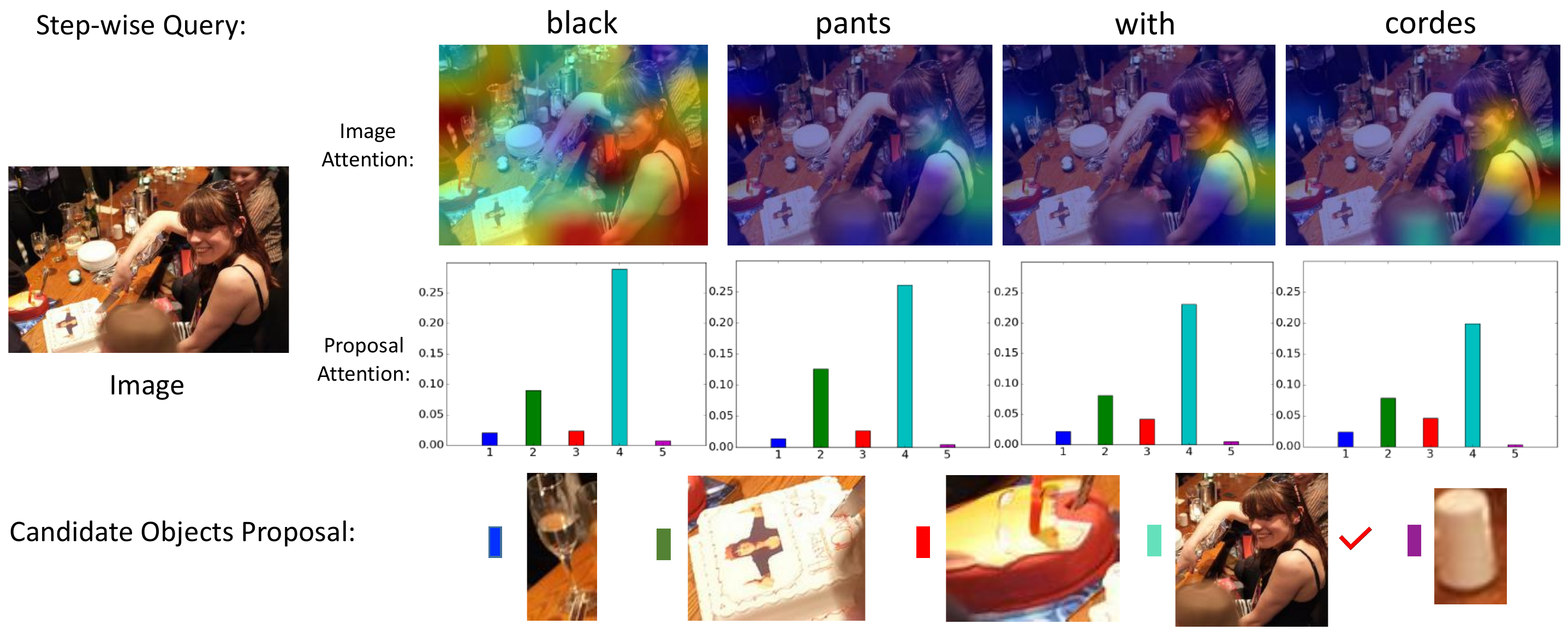} \\
		\end{tabular}
	}
	\vspace{-7pt}
	\caption{Qualitative results. We explicitly show the recurrent parallel attention change during inference to make the referring process explainable and visible. The referring queries are shown above the image-level attention maps. The histograms indicate the proposal-level attention on the top-5 proposals. The ground truth answer is marked. Notice the last row is a fail case.}
	\label{fig:exp_guesswhat}
\end{figure*}

Overall, RefCOCO has 142,210 expressions for 50,000 objects in 19,994 images. Compared to RefCOCO, RefCOCO+ removes absolute location words in referring expressions and contains 141,565 expressions for 49,856 objects in 19,992 images. Moreover, RefCOCOg has longer expressions and includes 104,560 expressions for 54,822 objects in 26,711 images. What's more, we use the ``unc'' standard splits for RefCOCO and RefCOCO+ datasets while using ``google'' split for RefCOCOg for evaluation. RefCOCO and RefCOCO+ provide person vs. object splits for evaluation. Images containing multiple people are in the `TestA' while images containing multiple objects are in the `TestB'. The
GuessWhat?! dataset~\cite{Vries_2017_CVPR} is a two-player guessing game where the questioner asks a series of questions and gives answers while the guesser predicts the correct object given the questioner's information. The dataset is composed of 155,280 dialogues containing 821,889 questions/answer pairs on 66,537 unique images and 134,073 unique objects. In this paper, our method deals with the guesser task proposed in the GuessWhat dataset. Given an image and a sequence of questions and answers, the task is to predict the correct object from the set of all object candidates. 

\subsection{Evaluation on RefCOCO, RefCOCO+ and RefCOCOg datasets} \label{sec:refcoco}
The object candidates can be obtained through a pre-trained object detection model like Faster-RCNN \cite{Ren2015Faster}, or an object proposal model such as Edgebox \cite{zitnick2014edge}, Objectness \cite{alexe2012measuring} and so on. However, for a fair comparison with previous methods, we follow \cite{Yu_2017_CVPR,yu2016modeling} use all the annotated entities in the image as the proposal bounding boxes at both training and test.

The results are reported in Table~\ref{tab:refcoco}. Compared to previous state-of-the-art methods, our methods have steadily improvements on all the three datasets. Interestingly, our simple baseline model is still competitive. It proves that by calculating the similarity between the local visual features and the encoded image and language features is a reasonable framework for grounding the referring target. It can be seen that by adding the image-level attention and the proposal-level attention separately, we further observe the performance increase. This shows that recurrently attending to the informative image regions/proposals according to the language descriptions can effectively filter the noise from unrelated regions/proposals for better encoding. The effect of the proposal-level attention is more obvious.

For our full model, the two-level attention models are jointly optimized and the results are further improved. It can be attributed to that the reasonable image-level attended region is consistent with the attended proposals, which further promotes the grounding of the target object. Note that the proposed method only focus on improving the model that is similar to the listener in the latest state-of-the-art~\cite{Yu_2017_CVPR}. In~\cite{Yu_2017_CVPR}, the speaker is a generative model that aims to produce referring expressions. The listener learns to embed the visual information and referring expression into a joint embedding space for comprehension and the reinforcer introduces a reward function to guide sampling of more discriminative expressions. By comparing our full model with the listener model of~\cite{Yu_2017_CVPR}, for instance, we outperform it by nearly 5\% on TestA setting of RefCOCO+. Our final single model is even better than a ensemble model in ~\cite{Yu_2017_CVPR}.

On the RefCOCOg, we outperforms the MMI \cite{nagaraja2016modeling} and other state-of-the-art on the same split. \cite{Yu_2017_CVPR} divide the dataset by randomly partitioning objects into training and validation splits while we use the standard `google' split for evaluation, which is not comparable.

\begin{table}[t]
	\centering
	\resizebox{0.70\linewidth}{!}{
		\begin{tabular}{l | c c}
			\hline
			Methods &val &test \\\hline
			Human &9.2\% &9.2\% \\
			Random &82.9\% &82.9\% \\\hline
			LSTM~\cite{Vries_2017_CVPR}  &37.9\%   &38.7\%  \\
			HRED~\cite{Vries_2017_CVPR}  &38.2\%  &39.0\%  \\
			LSTM+VGG~\cite{Vries_2017_CVPR} & 38.5\%  &39.5\%  \\
			HRED+VGG~\cite{Vries_2017_CVPR} &38.4\%   &39.6\%   \\\hline
			Baseline &37.9\% &39.0\% \\
			Image-level attention &37.4\%  &38.2\% \\
			Proposal-level attention &36.9\%  &37.1\% \\
			Parallel attention &\bf{36.2}\%  &\bf{36.6}\%  \\\hline
		\end{tabular}}
		\vspace{-3pt}
		\caption{Classification errors for the guesser on validation and test set.}
		\label{tab:guesswhat}
		\vspace{-3pt}
	\end{table}

\begin{table}[t!]
	\centering
	\resizebox{0.60\linewidth}{!}
	{
		\begin{tabular}{ l | c c}
			\hline
			Methods &val &test \\\hline
			LSTM~\cite{Vries_2017_CVPR} &50.9\% &51.4\%  \\
			Baseline &48.5\%  &46.4\% \\
			Parallel attention &\bf{44.3}\%  &\bf{40.3}\%  \\\hline
		\end{tabular}}
		\vspace{-3pt}
		\caption{Classification error rate for the guesser on validation and test set without using category features.}
		\label{tab:guesswhat_2}
		\vspace{-12pt}
	\end{table}

\subsection{Evaluation on the GuessWhat?! dataset}  \label{sec:guesswhat}
The results on the GuessWhat?! dataset are reported in Table~\ref{tab:guesswhat} and Table~\ref{tab:guesswhat_2}. We only compare our methods with the guesser model proposed in~\cite{Vries_2017_CVPR} since we share the same task.
In Table~\ref{tab:guesswhat}, we concatenate the object category feature with the final representation of the proposal-way framework for a fair comparison. As for the object category, we convert its one-hot class vector into a dense 512-dimensional category embedding using a learned look-up table. We can see steady improvement by comparing our approach with baseline methods in~\cite{Vries_2017_CVPR}. 
From the results, we can observe that the similar trend as shown in Sec~\ref{sec:refcoco}. 
First, the image-level and proposal-level parallel frameworks perform as a robust baseline in grounding referential expressions. Additionally, the attention mechanism on both sides contribute a lot to the final performance increase. 

During experiments, we find that the category feature has significant impact to the results. To eliminate its effect, we remove the category feature and separately report the results in Table~\ref{tab:guesswhat_2}. From the table, we can see that the performance gap becomes much larger compared to Table~\ref{tab:guesswhat}, which further proves the effectiveness of the proposed methods. Especially, the LSTM model (the best model in~\cite{Vries_2017_CVPR}) considers the dialogue as one flat sequence to encode it into an embedding vector. By comparing our recurrent encoding baseline model with it, we observe obvious performance increase (\eg 5.0\% relative increase on the test set).

\subsection{Recurrent parallel attention visualization}
We also qualitatively illustrate the attention evolution process along with the input expression query and dialog. The results are shown in Figure~\ref{fig:exp_guesswhat}. From the results, we can observe that the image-attention and proposal-way attention both change reasonably according to the input expression or dialog for reasoning correct target object. For example, in Figure~\ref{fig:exp_guesswhat}, when the questioner speaks a dialog "Is it a bowl? No", the guesser is confusing with which object is the questioner referring because of the limited information. However, when the guesser model listens to more clues, we can see the image-level attention scores on the target turns to be quite high finally while the proposal-level attention also select the first proposal as the target object. It explicitly shows that the guesser model progressively feel more confident to localize the first proposal as the referring target. 
And we also show a failure case which corresponds to the last row in Figure~\ref{fig:exp_guesswhat}. We can see that the attention score on the target proposal becomes smaller during steps going. Even though the image-way attention and the final result is right, the trend of proposal-way attention fails.

	 \section{Conclusion}
In this paper, we have proposed to solve the referring expression comprehension task using a novel parallel attention network. We have proposed to recurrently discover the object with variable-length language descriptions, from phrases to dialogs. To achieve this goal, we employ a two-way attention mechanism to localize the referring object on the global contextual features from the whole image and to select the referring proposal simultaneously. Since we use a recurrent style to discover the object, we make a step towards the model interpretability. With extensive experiments, we validate the advantage of our proposed methods and produce the state-of-the-art performance on several benchmarked datasets.
	 \clearpage
\small{
	\bibliographystyle{ieee}
	\bibliography{reference}
}
	
\end{document}